\documentclass[conference]{IEEEtran}

\usepackage{graphicx}
\usepackage{booktabs}
\usepackage{hyperref}
\usepackage{cite}
\usepackage{amsmath}
\usepackage{xcolor}
\usepackage{tikz}
\usepackage{float}
\usepackage[T1]{fontenc}
\usepackage[utf8]{inputenc}
\usepackage{enumitem}
\usepackage{caption}
\usepackage{subcaption}
\usepackage{url}
\usepackage{balance}
\usepackage{pifont}
\usepackage{multirow}

\usetikzlibrary{shapes.geometric, arrows.meta, positioning, fit, calc, backgrounds}

\hypersetup{
    colorlinks=true,
    linkcolor=blue!60!black,
    citecolor=green!50!black,
    urlcolor=blue!70!black,
}

\definecolor{layer1}{RGB}{31,78,121}
\definecolor{layer2}{RGB}{46,117,182}
\definecolor{layer3}{RGB}{91,155,213}
\definecolor{layer4}{RGB}{142,180,227}
\definecolor{crosscut}{RGB}{89,89,89}
\definecolor{citizen}{RGB}{68,114,148}

\title{GovAI-Pipe: A Layered AI Governance Pipeline for\\
Citizen-Facing AI in Turkey's e-Government Gateway}

\author{
  \IEEEauthorblockN{Ahmet Kaplan}
  \IEEEauthorblockA{
    Department of Computer Science\\
    Istanbul Medipol University\\
    Istanbul, T\"{u}rkiye\\
    ahmet.kaplan@medipol.edu.tr
  }
}

\begin{document}

\maketitle

\begin{abstract}
Turkey's e-Government Gateway (e-Devlet) serves over 68 million registered users with more than 9,200 government services, and is increasingly integrating artificial intelligence into citizen-facing applications such as chatbot assistants and eligibility assessments. However, no structured technical governance infrastructure currently connects high-level AI policy frameworks, such as the EU AI Act, OECD AI Principles, and Turkey's own National AI Strategy, to the operational reality of deploying AI within a centralized e-government platform. We propose GovAI-Pipe, a four-layer governance pipeline designed using Design Science Research methodology that maps the AI model lifecycle to governance checkpoints: (1)~pre-deployment validation for bias testing, explainability, and privacy impact assessment; (2)~deployment governance for risk-tier classification and approval workflows; (3)~runtime monitoring for drift detection, fairness tracking, and human-in-the-loop escalation; and (4)~post-incident governance for audit trails, rollback, and citizen redress. Each layer is anchored to specific provisions of the EU AI Act, the GDPR data protection framework, and the National AI Strategy. We demonstrate the framework through two high-risk e-Devlet use cases, showing how GovAI-Pipe operationalizes governance principles as auditable, technical pipeline components.
\end{abstract}


\section{Introduction}

Turkey's e-Government Gateway (turkiye.gov.tr), known domestically as e-Devlet, has grown into one of the world's largest centralized digital government platforms. With over 68 million registered users---representing approximately 96\% of the population over age 15---and more than 9,200 integrated services spanning healthcare, taxation, social welfare, and civil registration, the platform processes billions of transactions annually. As artificial intelligence capabilities advance, Turkey's public sector is actively integrating AI across this infrastructure: chatbot assistants guide citizens through complex bureaucratic procedures, and machine learning models pre-screen eligibility for social benefit programs. The United Nations E-Government Survey found that AI integration in national e-government strategies is accelerating globally, yet fewer than half of all countries have articulated how AI should be governed in public administration~\cite{un2024}.

Despite this momentum, there is no structured technical governance infrastructure to ensure that citizen-facing AI on e-Devlet operates ethically, transparently, and accountably. At the international level, the EU AI Act introduces risk classification with strict compliance obligations for high-risk systems in public administration~\cite{euaiact2024}, while the OECD AI Principles provide a normative blueprint around transparency, robustness, and accountability~\cite{oecd2024}. On the industry side, MLOps practices offer developed technical infrastructure for managing machine learning lifecycles~\cite{kreuzberger2023}, yet these systems target commercial contexts and lack government-specific governance gates: rights to explanation and appeal, public accountability, and citizen data protection constraints. Turkey has articulated ambitious AI governance goals through the National AI Strategy (UYZS) 2021--2025, the Draft AI Law (2024), the 2026 Annual Program, and the GDPR data protection law~\cite{uyzs2021,turkeyailaw2024,turkey2026program,kvkk2016}, yet none provides an engineering blueprint for technical operationalization within e-Devlet.

A review of the literature reveals a compound research gap. International AI governance frameworks operate at the policy level without technical pipeline operationalization---the ``principles-to-practice gap''~\cite{mittelstadt2019,morley2020}. Industry MLOps provides a proven technical foundation but is not adapted for government-specific requirements~\cite{kreuzberger2023}. The e-government AI literature is predominantly adoption-focused with limited work on technical governance infrastructure~\cite{zuiderwijk2021, un2024}. Turkey's policy documents articulate governance goals but lack concrete engineering specifications for e-Devlet. To the best of our knowledge, no existing work bridges these domains by proposing a technical governance pipeline for citizen-facing e-government AI.

To address this gap, we propose GovAI-Pipe, a four-layer governance pipeline for citizen-facing AI in Turkey's e-Government Gateway. Following Design Science Research methodology~\cite{peffers2007}, our work makes three contributions. First, we present a four-layer architecture that maps the AI model lifecycle to governance checkpoints---pre-deployment validation, deployment governance, runtime monitoring, and post-incident governance---with each layer producing auditable governance artifacts. Second, we develop a requirements mapping that traces international governance principles (EU AI Act, OECD) and Turkish regulatory requirements (GDPR, Draft AI Law) to concrete pipeline components and technical controls. Third, we provide an illustrative application of the pipeline to two high-risk e-Devlet use cases, demonstrating how GovAI-Pipe adapts governance intensity to different service contexts.

The remainder of this paper is organized as follows. Section~II reviews related work and describes our methodology. Section~III presents the GovAI-Pipe architecture. Section~IV demonstrates the pipeline through two e-Devlet use cases. Section~V discusses implications, limitations, and concludes with future work.

\section{Background and Methodology}
\label{sec:background}

\subsection{Related Work}
\label{subsec:related-work}

The past five years have produced a rich but predominantly policy-oriented body of AI governance work. The EU AI Act~\cite{euaiact2024} establishes a four-tier risk classification with strict compliance obligations for high-risk systems in public administration, becoming fully applicable in August 2026. The OECD AI Principles~\cite{oecd2024}, endorsed by 47 jurisdictions including Turkey, provide a normative blueprint around transparency, robustness, and accountability. The UNESCO Recommendation on AI Ethics~\cite{unesco2021} anchors governance in human-rights-centered principles. At the national level, Singapore's Model AI Governance Framework~\cite{singapore2024} operationalizes governance through nine dimensions supported by the AI Verify testing toolkit, while the NIST AI Risk Management Framework~\cite{nist2023} structures governance around four core functions. Beneath these regulatory instruments lies a recurring scholarly diagnosis: the ``principles-to-practice gap.'' Jobin et al.~\cite{jobin2019} found global convergence around five ethical principles but substantial divergence in implementation. Mittelstadt~\cite{mittelstadt2019} argued that principles alone cannot guarantee ethical AI, and Morley et al.~\cite{morley2020} mapped AI ethics tools to ML pipeline stages, finding accountability and safety underserved. These frameworks define \emph{what} governance should achieve but not \emph{how} to implement these requirements as automated, pipeline-integrated technical controls.

MLOps provides the technical substrate upon which governance pipelines must be built. Kreuzberger et al.~\cite{kreuzberger2023} provided a consolidated definition of MLOps with a reference architecture featuring feature stores, model registries, metadata stores, and orchestrators. The foundational motivation traces to Sculley et al.~\cite{sculley2015}, who demonstrated that ML systems accumulate massive technical debt beyond model code. Fairness tools such as AIF360~\cite{bellamy2019} and Fairlearn~\cite{bird2020} translate theoretical fairness definitions into deployable components, while explainability frameworks including SHAP~\cite{lundberg2017} and LIME~\cite{ribeiro2016} enable per-decision interpretability. Model cards~\cite{mitchell2019} and datasheets for datasets~\cite{gebru2021} formalize accountability artifacts. Despite this technical depth, these tools target commercial contexts and are not adapted for government-specific requirements: citizen rights to explanation, public accountability to oversight bodies, and fundamental rights impact assessment.

AI adoption in government is accelerating but remains under-governed. The OECD~\cite{oecd2025gov} documented over 200 government AI examples, finding adoption trailing the private sector. The UN E-Government Survey~\cite{un2024} found that only 21\% of countries address ethical AI use in public administration. Veale and Brass~\cite{veale2019} called for transparency and public accountability as safeguards for algorithmic governance. Algorithmic impact assessments have emerged as the primary pre-deployment governance mechanism, with Canada's AIA Tool~\cite{canada2019} being the most operationally developed example. Country cases such as Estonia's X-Road~\cite{saputro2020}, Singapore's AI Verify, and the UK's Algorithmic Transparency Recording Standard~\cite{ukgds2025} demonstrate progressive but incomplete approaches---none offers a replicable technical pipeline architecture.

Turkey's e-Devlet platform serves over 68 million users with 9,200+ services from over 1,000 institutions~\cite{akan2025,oecd2023turkey}, coordinated by the Digital Transformation Office (CBDDO). Earlier work on improving citizen take-up of e-government services in Turkey identified practical barriers that remain relevant as AI-driven services raise new governance expectations~\cite{medeni2012}. Prior work on citizen-centric evaluation of Turkish e-government services~\cite{tsohou2013} established that structured evaluation frameworks improve service quality---a principle GovAI-Pipe extends to AI governance. The National AI Strategy (UYZS) 2021--2025~\cite{uyzs2021} defines six strategic priorities including a ``Trusted AI Seal'' for public-sector AI. The Draft AI Law~\cite{turkeyailaw2024} introduces EU AI Act--aligned risk-tier classification. The GDPR~\cite{kvkk2016} governs AI-driven personal data processing, with recent guidance on generative and agentic AI~\cite{kvkk2025}. The 2026 Annual Program~\cite{turkey2026program} positions AI as a governance layer with standardized certification. Despite this broad policy effort, Turkey has articulated \emph{what} governance outcomes are desired but lacks an engineering blueprint for \emph{how} to operationalize them within e-Devlet.

To the best of our knowledge, no existing work proposes a technical governance pipeline that bridges international AI principles with a specific country's regulatory requirements for citizen-facing e-government AI. GovAI-Pipe addresses this compound gap.

\subsection{Design Science Research}
\label{subsec:dsr}

We adopt Design Science Research (DSR) as our methodological foundation~\cite{hevner2004}. DSR focuses on creating and evaluating purposeful IT artifacts intended to solve identified organizational problems, and is the most common methodological choice in e-government research proposing technical artifacts~\cite{carter2022}. We select DSR because GovAI-Pipe is a prescriptive artifact that does not yet exist in practice and cannot be studied through purely observational methods.

Following the six-stage DSR process model of Peffers et al.~\cite{peffers2007}, we: (1)~identify the gap between Turkey's AI governance ambitions and the absence of a technical mechanism to operationalize them within e-Devlet; (2)~define the objective as a layered governance pipeline mapping regulatory requirements to technical checkpoints across the AI lifecycle; (3)~design GovAI-Pipe as a four-layer architecture derived from systematic mapping of governance principles to pipeline components; (4)~demonstrate applicability through two illustrative high-risk e-Devlet use cases; (5)~evaluate analytically using completeness, consistency, applicability, and comparison criteria appropriate for a first-iteration artifact; and (6)~communicate the artifact through this paper. Empirical validation through expert panels and prototype implementation is planned as future work.

\section{GovAI-Pipe Architecture}
\label{sec:architecture}

\subsection{Architecture Overview}
\label{subsec:architecture-overview}

%
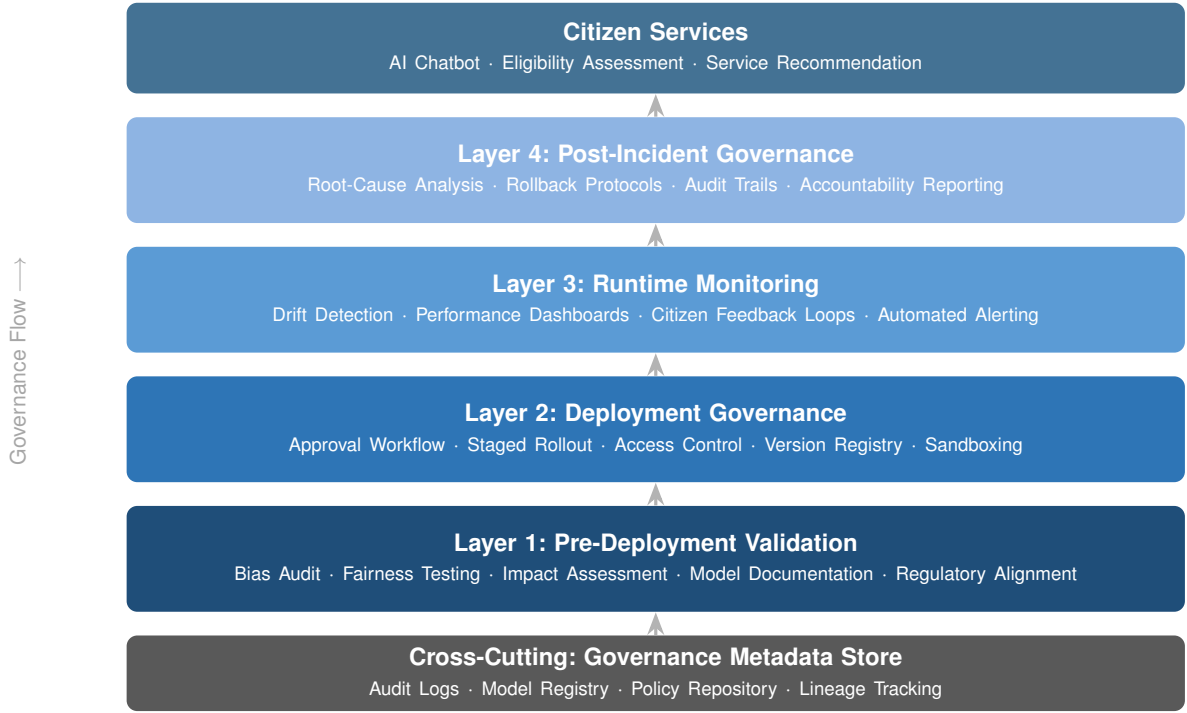
\begin{figure*}[t]
\centering
\begin{tikzpicture}[
    layerbox/.style={
        rectangle, rounded corners=4pt, minimum width=14cm, minimum height=1.4cm,
        text=white, font=\sffamily\small, align=center, text width=13cm,
        draw=none,
    },
    component/.style={
        font=\sffamily\scriptsize, text=white,
    },
    arrowstyle/.style={
        -{Stealth[length=3mm]}, line width=1.2pt, color=gray!60,
    },
    node distance=0.3cm,
]

\node[layerbox, fill=crosscut, minimum height=1.0cm] (crosscut)
    {\textbf{Cross-Cutting: Governance Metadata Store}\\[1pt]
    {\scriptsize Audit Logs \(\cdot\) Model Registry \(\cdot\) Policy Repository \(\cdot\) Lineage Tracking}};

\node[layerbox, fill=layer1, above=of crosscut] (L1)
    {\textbf{Layer 1: Pre-Deployment Validation}\\[1pt]
    {\scriptsize Bias Audit \(\cdot\) Fairness Testing \(\cdot\) Impact Assessment \(\cdot\) Model Documentation \(\cdot\) Regulatory Alignment}};

\node[layerbox, fill=layer2, above=of L1] (L2)
    {\textbf{Layer 2: Deployment Governance}\\[1pt]
    {\scriptsize Approval Workflow \(\cdot\) Staged Rollout \(\cdot\) Access Control \(\cdot\) Version Registry \(\cdot\) Sandboxing}};

\node[layerbox, fill=layer3, above=of L2] (L3)
    {\textbf{Layer 3: Runtime Monitoring}\\[1pt]
    {\scriptsize Drift Detection \(\cdot\) Performance Dashboards \(\cdot\) Citizen Feedback Loops \(\cdot\) Automated Alerting}};

\node[layerbox, fill=layer4, above=of L3] (L4)
    {\textbf{Layer 4: Post-Incident Governance}\\[1pt]
    {\scriptsize Root-Cause Analysis \(\cdot\) Rollback Protocols \(\cdot\) Audit Trails \(\cdot\) Accountability Reporting}};

\node[layerbox, fill=citizen, minimum height=1.2cm, above=of L4] (citizen)
    {\textbf{Citizen Services}\\[1pt]
    {\scriptsize AI Chatbot \(\cdot\) Eligibility Assessment \(\cdot\) Service Recommendation}};

\draw[arrowstyle] (crosscut.north) -- (L1.south);
\draw[arrowstyle] (L1.north) -- (L2.south);
\draw[arrowstyle] (L2.north) -- (L3.south);
\draw[arrowstyle] (L3.north) -- (L4.south);
\draw[arrowstyle] (L4.north) -- (citizen.south);

\node[rotate=90, font=\sffamily\footnotesize, text=gray!70, anchor=south]
    at ($(crosscut.west)!0.5!(citizen.west) + (-1.2,0)$)
    {Governance Flow \(\longrightarrow\)};

\end{tikzpicture}
\caption{GovAI-Pipe four-layer AI governance architecture for citizen-facing services on Turkey's e-Government Gateway (e-Devlet).}
\label{fig:govai-pipe-architecture}
\end{figure*}

GovAI-Pipe is organized as a four-layer governance pipeline through which every AI model must pass sequentially before serving citizens on Turkey's e-Government Gateway (Figure~\ref{fig:govai-pipe-architecture}). The four layers---pre-deployment validation, deployment governance, runtime monitoring, and post-incident governance---correspond to distinct phases of the AI model lifecycle and ensure that governance controls are applied continuously rather than as one-time compliance checks. This lifecycle-aligned design follows a lesson from the MLOps literature~\cite{kreuzberger2023}: governance bolted on after deployment is both fragile and incomplete.

The bottom-up flow is deliberate: a candidate model enters at Layer~1, undergoes validation before any deployment decision; Layer~2 determines whether and how the model may be released; Layer~3 monitors behavior in production with escalation paths; and Layer~4 activates when incidents occur. A cross-cutting governance metadata store underpins all layers, maintaining a policy-to-requirement traceability matrix, centralized audit logs, and a governance dashboard accessible to oversight bodies including Turkey's National Technology and AI Directorate and the CBDDO.

\subsection{Pipeline Layers}
\label{subsec:pipeline-layers}

\textbf{Layer 1: Pre-Deployment Validation.}
This layer is the first governance gate, comprising five components. The bias testing suite evaluates candidate models across demographic dimensions relevant to Turkey---gender, age, geographic region (with attention to East--West disparities), and disability status---computing standard fairness metrics (demographic parity, equalized odds, disparate impact ratio) against configurable thresholds aligned with the OECD AI Principle on non-discrimination~\cite{oecd2024} and EU AI Act high-risk system requirements~\cite{euaiact2024}. The explainability validator applies SHAP and LIME methods and evaluates explanation quality against coverage and faithfulness criteria, implementing EU AI Act Article~13 transparency requirements. The privacy impact assessment (PIA) engine checks compliance with GDPR Articles~4--6 for models processing citizen personal data~\cite{kvkk2016}. The dataset documentation component generates structured documentation following the datasheets methodology~\cite{gebru2021}. Finally, each model receives a model card~\cite{mitchell2019} that acts as a ``governance passport'' through the remaining layers.

\textbf{Layer 2: Deployment Governance.}
This layer controls the transition from validated model to production service. A centralized model registry records version history, ownership, metadata, and governance artifacts, operationalizing Turkey's Draft AI Law registry requirements~\cite{turkeyailaw2024}. A multi-stage approval workflow implements a structured chain: development team self-certification, independent governance team review, and CBDDO final authorization, with high-risk models additionally requiring sectoral ministry sign-off---implementing EU AI Act Article~14 human oversight~\cite{euaiact2024}. Canary and shadow deployment infrastructure mitigates rollout risk by serving limited traffic before full release. A GDPR compliance gate verifies data residency within Turkey and consent requirements for sensitive personal data (GDPR Article~6). The EU AI Act risk-tier classification gate assigns each model to one of four risk tiers, determining governance intensity throughout the remaining pipeline.

\textbf{Layer 3: Runtime Monitoring.}
This layer provides continuous oversight of deployed models. Drift detection continuously computes distributional distance metrics (PSI, KL divergence) between production and training data, triggering alerts and automatic traffic reduction for high-risk models when thresholds are exceeded---operationalizing the EU AI Act Article~15 robustness requirement~\cite{euaiact2024}. Fairness metric tracking monitors the same demographic metrics evaluated at Layer~1 on live production data, with degradation triggering escalation. Explainability health checks periodically recompute explanations on production samples to verify they remain above established thresholds. SLA monitoring tracks latency, availability, and error rates, which acquire governance significance when service failures deny citizens access to government services. All alerts feed into a human-in-the-loop escalation engine that routes issues to appropriate decision-makers, with high-risk models prohibiting fully autonomous resolution.

\textbf{Layer 4: Post-Incident Governance.}
This layer addresses failures, bias events, and citizen complaints. Automated rollback triggers revert affected models to the last known-good version upon critical governance failures, with immediate rollback for high-risk services. The impact assessment engine quantifies scope and severity of harm, identifying affected citizens and estimating downstream consequences. An immutable audit trail records every governance-relevant event across all layers, providing the evidence chain for accountability under both the EU AI Act Article~17~\cite{euaiact2024} and Turkey's 2026 Annual Program~\cite{turkey2026program}. The citizen grievance mechanism operationalizes GDPR Article~11 rights to object to automated decisions~\cite{kvkk2016}, providing structured challenge and explanation pathways. Grievance patterns feed back into the model improvement cycle, triggering revalidation at Layer~1 when systematic issues emerge.

\subsection{Governance Requirements Mapping}
\label{subsec:requirements-mapping}

Table~\ref{tab:requirements-mapping} presents the mapping from governance principles to technical pipeline components, providing an auditable traceability chain from policy requirements to enforcement mechanisms.

\begin{table*}[t]
\centering
\caption{Governance requirements mapping: from policy principles to technical pipeline components.}
\label{tab:requirements-mapping}
\small
\begin{tabular}{@{}p{2.2cm}p{2.8cm}p{2.0cm}p{3.2cm}p{3.8cm}@{}}
\toprule
\textbf{Governance Principle} & \textbf{Source} & \textbf{Pipeline Layer} & \textbf{Technical Component} & \textbf{Metric / Artifact} \\
\midrule
Risk classification & EU AI Act Art.~6--7; Draft AI Law & L2: Deployment & Risk-tier classification gate & Risk level (Unacceptable / High / Limited / Minimal) \\
\addlinespace
Bias \& non-discrimination & OECD AI 1.2; EU AI Act Art.~10 & L1: Pre-deploy & Bias testing suite & Demographic parity, equalized odds, disparate impact ratio \\
\addlinespace
Transparency & OECD AI 1.3; EU AI Act Art.~13 & L1: Pre-deploy & Model card generator; explainability validator & SHAP coverage \%; explanation fidelity score \\
\addlinespace
Human oversight & EU AI Act Art.~14 & L2: Deploy; L3: Runtime & Approval workflow; HITL escalation & Approval latency; escalation rate \\
\addlinespace
Data governance & GDPR Art.~4--6 & L1: Pre-deploy; L2: Deploy & PIA automation; data residency check & PIA pass/fail; data location verification \\
\addlinespace
Robustness \& safety & OECD AI 1.4; EU AI Act Art.~15 & L3: Runtime & Drift detection; SLA monitoring & PSI / KL divergence; error rate; uptime \\
\addlinespace
Accountability & EU AI Act Art.~17; UYZS & L4: Post-incident & Immutable audit trail; impact assessment & Trace completeness \%; incident response time \\
\addlinespace
Citizen redress & EU AI Act Art.~68; GDPR Art.~11 & L4: Post-incident & Grievance feedback loop & Complaint resolution rate; model update trigger rate \\
\bottomrule
\end{tabular}
\end{table*}

The mapping demonstrates three distinguishing properties: completeness---all eight governance principles are mapped to at least one pipeline layer and technical component; traceability---any governance incident can be traced from a technical artifact to its originating policy requirement; and measurability---each principle is associated with quantifiable metrics, enabling objective compliance assessment rather than self-reported adherence to abstract principles.

\section{Illustrative Application}
\label{sec:application}

We demonstrate GovAI-Pipe through two HIGH-risk citizen-facing e-Devlet use cases that exercise all four pipeline layers. LIMITED-risk use cases (e.g., personalized service recommendation engines) are deferred to future work.

\subsection{UC1: AI Chatbot Assistant}
\label{subsec:uc-chatbot}

The first use case concerns a Turkish-language AI chatbot (e-Devlet Asistan) deployed on turkiye.gov.tr to help citizens navigate over 9,200 government services. Given the platform's 68 million users and the risk that erroneous guidance could lead citizens to miss deadlines or forfeit benefits, the chatbot requires strict governance.

Layer~1 subjects the chatbot to bias testing across Turkish regional dialects (ensuring citizens from southeastern Anatolia receive equivalent quality to those using standard Istanbul Turkish), gender bias audits, and age-related accessibility testing. Explainability validation ensures responses link to specific government regulations. Artifacts include a bias audit report with pass/fail determinations against predefined thresholds (e.g., demographic parity difference $< 0.05$) and a model card~\cite{mitchell2019} documenting training data provenance, intended use boundaries, and performance characteristics.

Layer~2 assigns a HIGH-risk designation given the volume of affected citizens and potential influence on administrative rights. This triggers multi-stage approval (NLP team, data governance unit, CBDDO final authorization). GDPR compliance checks verify that explicit informed consent is obtained before retaining conversation data containing personal information. Deployment proceeds through canary rollout at 5\% traffic. Artifacts include a risk-tier certificate, approval chain log, and GDPR consent verification record.

Layer~3 continuously monitors hallucination rates (fabricated procedural guidance not grounded in official sources), response accuracy drift as regulations change, and demographic fairness in response quality and resolution rates. When metrics cross thresholds---e.g., hallucination rate exceeding 2\%---automated alerts escalate to the human-in-the-loop review queue. A real-time governance dashboard is accessible to the CBDDO oversight team.

Layer~4 activates when citizens report incorrect guidance. Complaints are cross-referenced against conversation logs; substantiated complaints generate an incident report with root cause analysis. Severe errors trigger automated rollback. Aggregated complaint patterns feed into retraining, closing the governance loop.

\subsection{UC2: Social Benefit Eligibility Assessment}
\label{subsec:uc-eligibility}

The second use case involves an AI system that pre-screens citizen eligibility for social welfare programs (disability benefits, housing assistance, educational scholarships). The model analyzes income records, household composition, geographic location, and disability status to produce preliminary eligibility recommendations for human caseworker review.

Layer~1 conducts thorough fairness testing given Turkey's socioeconomic disparities. The bias testing suite evaluates performance disaggregated by geographic region (eastern provinces versus western metropolitan areas), income level, gender, and disability status, with a minimum disparate impact threshold of 0.80 (the four-fifths rule). The system also validates eligibility explanations for legal sufficiency under GDPR Article~11, as citizens have a right to understand why they were deemed ineligible. Artifacts include a disparate impact report and validated explanation templates.

Layer~2 assigns HIGH-risk classification given direct impact on fundamental socioeconomic rights, triggering the most stringent controls. A mandatory human-in-the-loop configuration is enforced: the AI may only recommend outcomes, with a qualified caseworker approving every determination. Because the system processes GDPR Article~6 sensitive data (health and financial records), approval requires sign-off from both the Ministry of Family and Social Services and the GDPR board. Artifacts include a HITL workflow configuration document and sensitive data processing certificate.

Layer~3 tracks approval and rejection rates disaggregated by all demographic dimensions, with alerts triggered when subgroup rates deviate beyond two standard deviations. Socioeconomic drift detection monitors shifts in applicant population distribution---for instance, regional economic downturns causing surges from previously underrepresented areas. A fairness dashboard provides the oversight authority with longitudinal equity metrics.

Layer~4 operationalizes GDPR Article~11 objection rights. When a citizen objects, the system retrieves the complete audit trail and produces a decision trace document reconstructing the full reasoning chain. Objections trigger mandatory re-review by an independent senior caseworker, and if systematic bias is detected, a model suspension and retraining cycle is initiated.

\subsection{Cross-Case Observations}

Three observations emerge from the two use cases. First, risk-tier differentiation: both HIGH-risk cases trigger multi-authority approvals and mandatory HITL, demonstrating the pipeline's ability to enforce proportionate governance intensity. Second, Turkey-specific elements distinguish GovAI-Pipe from generic frameworks: Turkish dialect bias testing, East--West regional fairness, GDPR Article~11 objection workflows, and CBDDO as centralized approval authority. Third, engineering concreteness: every layer produces auditable artifacts (bias reports, risk certificates, fairness dashboards, decision traces), ensuring governance produces durable evidentiary records. LIMITED-risk use cases, such as personalized service recommendation engines, would trigger streamlined approval and lighter monitoring---a differentiation we plan to elaborate in an extended version.

\section{Discussion and Conclusion}
\label{sec:discussion-conclusion}

\subsection{Framework Comparison}

Table~\ref{tab:framework-comparison} situates GovAI-Pipe among existing AI governance and operational frameworks across ten governance dimensions.

\begin{table*}[t]
\centering
\caption{Feature comparison of GovAI-Pipe with existing governance and operational frameworks. \ding{51} = fully addressed; $\circ$ = partially addressed; \ding{55} = not addressed.}
\label{tab:framework-comparison}
\small
\begin{tabular}{lcccc}
\hline
\textbf{Governance Feature} & \textbf{GovAI-Pipe} & \textbf{EU AI Act Toolbox} & \textbf{Industry MLOps} & \textbf{Existing eGov Frameworks} \\
\hline
Risk classification (tiered)         & \ding{51} & \ding{51} & \ding{55} & $\circ$ \\
Bias \& fairness testing             & \ding{51} & $\circ$   & $\circ$   & \ding{55} \\
Explainability validation            & \ding{51} & $\circ$   & $\circ$   & \ding{55} \\
Human oversight \& escalation        & \ding{51} & \ding{51} & \ding{55} & $\circ$ \\
Immutable audit trail                & \ding{51} & $\circ$   & \ding{51} & \ding{55} \\
Citizen redress mechanism            & \ding{51} & $\circ$   & \ding{55} & $\circ$ \\
Country-specific compliance          & \ding{51} & \ding{55} & \ding{55} & $\circ$ \\
Deployment governance gates          & \ding{51} & $\circ$   & \ding{51} & \ding{55} \\
Runtime drift \& performance monitoring & \ding{51} & \ding{55} & \ding{51} & \ding{55} \\
Post-incident response protocol      & \ding{51} & $\circ$   & $\circ$   & \ding{55} \\
\hline
\end{tabular}
\end{table*}

The EU AI Act compliance toolbox provides strong normative foundations but is a regulatory instrument rather than a technical pipeline. Industry MLOps platforms excel at deployment and monitoring but lack government-specific controls such as citizen redress and fundamental rights impact assessment~\cite{kreuzberger2023}. Existing e-government frameworks---Estonia's X-Road, Singapore's AI Verify, the UK's Algorithmic Transparency Recording Standard---address selected dimensions but none provides a full, layered pipeline spanning the AI lifecycle. GovAI-Pipe's contribution is integrating regulatory requirements as automated pipeline gates, MLOps-grade infrastructure, and government-specific controls (citizen grievance-to-model feedback loops, GDPR compliance, CBDDO workflows) into a unified four-layer architecture.

Although designed for Turkey's e-Devlet, GovAI-Pipe's architecture is intentionally modular. Generic components (model registry, fairness testing, drift detection, audit trail) can be adopted with minimal modification by any centralized e-government platform. Jurisdiction-specific components (GDPR checks, CBDDO workflows, Turkish-language testing) would be replaced for other national contexts while retaining the four-layer governance sequencing.

\subsection{Limitations}

We acknowledge four limitations. First, GovAI-Pipe is a conceptual framework evaluated analytically; it has not been empirically validated through implementation or expert evaluation. A prototype implementation and Delphi study with government IT practitioners constitute the most important next steps. Second, the use cases are illustrative rather than empirical, constructed from publicly available information rather than actual deployed systems. Third, Turkey's regulatory environment is evolving: the Draft AI Law may undergo amendments before enactment, and the GDPR's recent agentic AI guidance is still being interpreted~\cite{turkeyailaw2024,kvkk2025}. Fourth, the framework assumes a centralized e-government architecture where a single authority can enforce compliance, which may not transfer directly to federated digital government structures.

\subsection{Conclusion and Future Work}

We proposed GovAI-Pipe, a four-layer governance pipeline embedding governance checkpoints across the AI model lifecycle for citizen-facing services on Turkey's e-Government Gateway. The pipeline covers pre-deployment validation (bias testing, explainability, PIA), deployment governance (risk-tier classification, approval workflows, GDPR compliance), runtime monitoring (drift detection, fairness tracking, HITL escalation), and post-incident governance (audit trails, rollback, citizen redress). A governance requirements mapping traces international and Turkish regulatory principles to concrete technical controls, and two illustrative HIGH-risk use cases demonstrate the pipeline's applicability and risk-tier differentiation.

Three directions for future work follow. First, we plan an expert validation through a Delphi study with Turkish government IT officials and data protection specialists. Second, we intend to develop a prototype implementation integrating open-source fairness toolkits, model registries, and monitoring frameworks for one e-Devlet use case. Third, we aim to investigate cross-country adaptation by applying GovAI-Pipe to other e-government platforms pursuing EU AI Act alignment. As AI becomes a standard layer of public governance, structured and auditable governance pipelines will be necessary; GovAI-Pipe offers a replicable blueprint for governments seeking to deploy AI responsibly for their citizens.

\balance
\bibliographystyle{ieeetr}
\bibliography{references}

\end{document}